\def\year{2017}\relax
\documentclass[letterpaper]{article} 
\usepackage{aaai17}  
\usepackage{times}  
\usepackage{helvet}  
\usepackage{courier}  
\usepackage{url}  
\usepackage{graphicx}  

\usepackage{amsmath}
\usepackage{array}
\usepackage[linesnumbered,ruled]{algorithm2e}
\usepackage{tabularx}
\usepackage{amssymb}

\usepackage{lipsum}
\usepackage{subcaption}

\newtheorem{prop}{Proposition}
\newcolumntype{L}[1]{>{\raggedright\let\newline\\\arraybackslash\hspace{0pt}}m{#1}}
\newcolumntype{C}[1]{>{\centering\let\newline\\\arraybackslash\hspace{0pt}}m{#1}}
\newcolumntype{R}[1]{>{\raggedleft\let\newline\\\arraybackslash\hspace{0pt}}m{#1}}

\begin{document}
%
\title{Adversarial Dropout for Supervised and Semi-Supervised Learning}
\author{Sungrae Park \and Jun-Keon Park \and Su-Jin Shin \and Il-Chul Moon\\
  Department of Industrial and System Engineering\\
  KAIST\\
  Deajeon, South Korea \\
  \{ \texttt{sungraepark} \and \texttt{alex3012} \and \texttt{sujin.shin} \and \texttt{icmoon} \} \texttt{@kaist.ac.kr}
}
\maketitle

\frenchspacing

\begin{abstract} 
Recently, training with adversarial examples, which are generated by adding a small but worst-case perturbation on input examples, has improved the generalization performance of neural networks. In contrast to the biased individual inputs to enhance the generality, this paper introduces \emph{adversarial dropout}, which is a minimal set of dropouts that maximize the divergence between 1) the training supervision and 2) the outputs from the network with the dropouts. The identified adversarial dropouts are used to automatically reconfigure the neural network in the training process, and we demonstrated that the simultaneous training on the original and the reconfigured network improves the generalization performance of supervised and semi-supervised learning tasks on MNIST, SVHN, and CIFAR-10. We analyzed the trained model to find the performance improvement reasons. We found that adversarial dropout increases the sparsity of neural networks more than the standard dropout. Finally, we also proved that adversarial dropout is a regularization term with a rank-valued hyper parameter that is different from a continuous-valued parameter to specify the strength of the regularization.
\end{abstract}

\section{Introduction}
\noindent Deep neural networks (DNNs) have demonstrated the significant improvement on benchmark performances in a wide range of applications. As neural networks become deeper, the model complexity also increases quickly, and this complexity leads DNNs to potentially overfit a training data set. Several techniques \cite{hinton2012improving,poole2014analyzing,bishop1995regularization,lasserre2006principled} have emerged over the past years to address this challenge, and \emph{dropout} has become one of dominant methods due to its simplicity and effectiveness \cite{hinton2012improving,srivastava2014dropout}. 


\emph{Dropout} randomly disconnects neural units during training as a method to prevent the feature co-adaptation \cite{baldi2013understanding,wager2013dropout,wang2013fast,li2016improved}. The earlier work by Hinton et al. \shortcite{hinton2012improving} and Srivastava et al. \shortcite{srivastava2014dropout} interpreted dropout as an extreme form of model combinations, a.k.a. a model ensemble, by sharing extensive parameters on neural networks. They proposed learning the model combination through minimizing an expected loss of models perturbed by dropout. They also pointed out that the output of dropout is the geometric mean of the outputs from the model ensemble with the shared parameters. Extending the weight sharing perspective, several studies  \cite{baldi2013understanding,chen2014dropout,jain2015drop} analyzed the ensemble effects from the dropout.

The recent work by Laine \& Aila (\citeyear{laine2016temporal}) enhanced the ensemble effect of dropout by adding self-ensembling terms. The self-ensembling term is constructed by a divergence between two sampled neural networks from the dropout. By minimizing the divergence, the sampled networks learn from each other, and this practice is similar to the working mechanism of the ladder network \cite{rasmus2015semi}, which builds a connection between an unsupervised and a supervised neural network. Our method also follows the principles of self-ensembling, but we apply the adversarial training concept to the sampling of neural network structures through dropout.

At the same time that the community has developed the dropout, \emph{adversarial training} has become another focus of the community. Szegedy et al. (\citeyear{szegedy2013intriguing}) showed that a certain neural network is vulnerable to a very small perturbation in the training data set if the noise direction is sensitive to the models' label assignment $y$ given $x$, even when the perturbation is so small that human eyes cannot discern the difference. They empirically proved that robustly training models against adversarial perturbation is effective in reducing test errors. However, their method of identifying adversarial perturbations contains a computationally expensive inner loop. To compensate it, Goodfellow et al. (\citeyear{goodfellow2014explaining}) suggested an approximation method, through the linearization of the loss function, that is free from the loop. Adversarial training can be conducted on supervised learning because the adversarial direction can be defined when true $y$ labels are known. Miyato et al. (\citeyear{miyato2015distributional}) proposed a virtual adversarial direction to apply the adversarial training in the semi-supervised learning that may not assume the true $y$ value. Until now, the adversarial perturbation can be defined as a unit vector of additive noise imposed on the input or the embedding spaces \cite{szegedy2013intriguing,goodfellow2014explaining,miyato2015distributional}.

Our proposed method, \emph{adversarial dropout}, can be viewed from the \emph{dropout} and from the \emph{adversarial training} perspectives. Adversarial dropout can be interpreted as dropout masks whose direction is optimized \textit{adversarially} to the model's label assignment. However, it should be noted that adversarial dropout and traditional adversarial training with additive perturbation are different because adversarial dropout induces the sparse structure of neural network while the other does not make changes in the structure of the neural network, directly. 
\begin{figure}[t]
  \centering
  	\includegraphics[width=8cm]{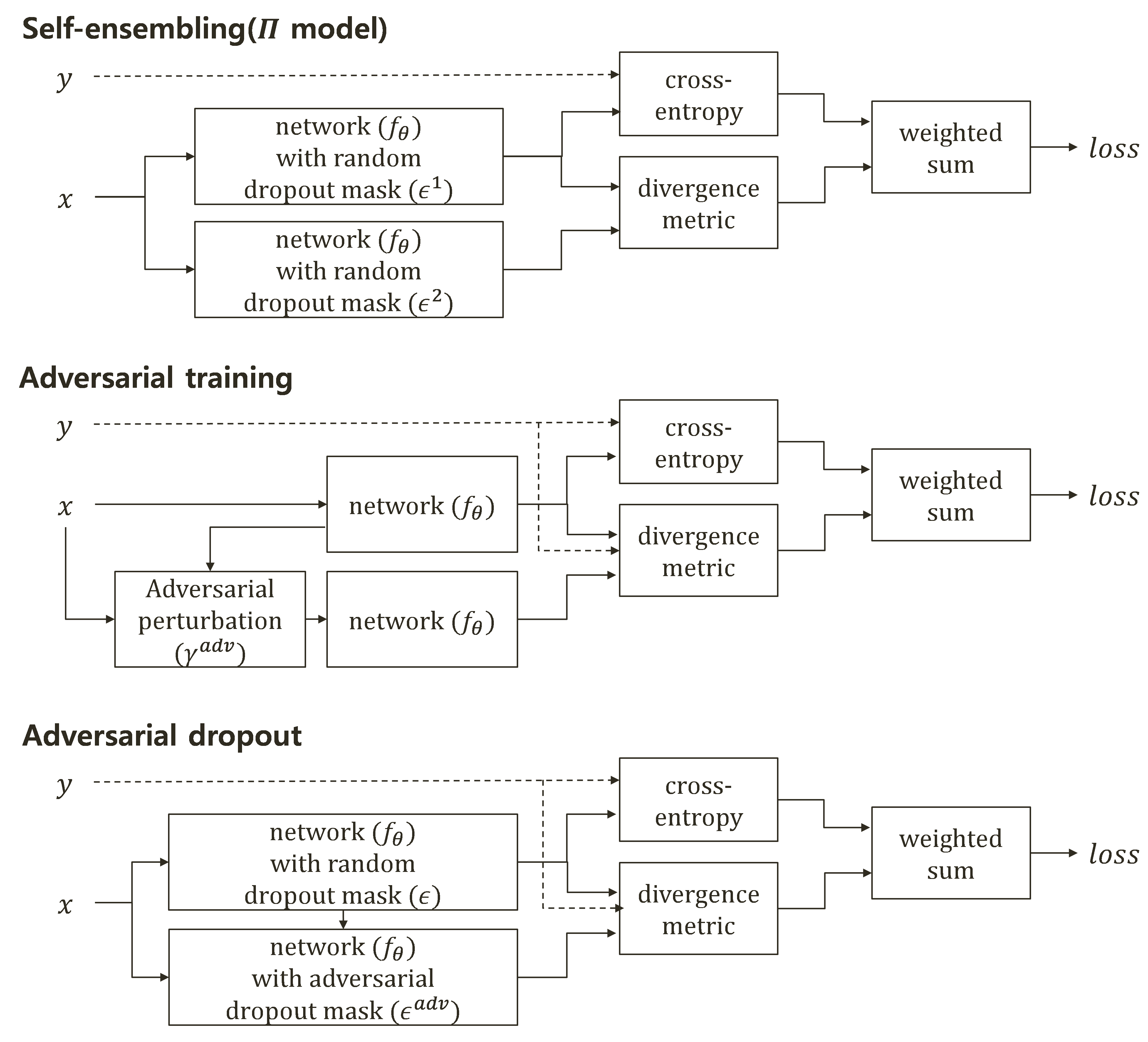}
  \caption{Diagram description of loss functions from $\Pi$ model \cite{laine2016temporal}, the adversarial training \cite{miyato2015distributional}, and our adversarial dropout. } 
\end{figure}

Figure 1 describes the proposed loss function construction of adversarial dropout compared to 1) the recent dropout model, which is $\Pi$ model \cite{laine2016temporal} and 2) the adversarial training \cite{goodfellow2014explaining,miyato2015distributional}. 
When we compare adversarial dropout to $\Pi$ model, both divergence terms are similarly computed from two different dropped networks, but adversarial dropout uses an optimized dropped network to adapt the concept of adversarial training. When we compare adversarial dropout to the adversarial training, the divergence term of the adversarial training is computed from one network structure with two training examples: clean and adversarial examples. On the contrary, the divergence term of the adversarial dropout is defined with two network structures: a randomly dropped network and an adversarially dropped network.   

Our experiments demonstrated that 1) adversarial dropout improves the performance on MNIST supervised learning compared to the dropout suggested by $\Pi$ model, and 2) adversarial dropout showed the state-of-the-art performance on the semi-supervised learning task on SVHN and CIFAR-10 when we compare the most recent techniques of dropout and adversarial training. Following the performance comparison, we visualize the neural network structure from adversarial dropout to illustrate that the adversarial dropout enables a sparse structure compared to the neural network of standard dropout. Finally, we theoretically show the original characteristics of adversarial dropout that specifies the strength of the regularization effect by the rank-valued parameter while the adversarial training specifies the strength with the conventional continuous-valued scale. 

\section{Preliminaries}

Before introducing adversarial dropout, we briefly introduce stochastic noise layers for deep neural networks. Afterwards, we review adversarial training and temporal ensembling, or $\Pi$ model, because two methods are closely related to adversarial dropout.

\subsection{Noise Layers}
Corrupting training data with noises has been well-known to be a method to stabilize prediction \cite{bishop1995training,maaten2013learning,wager2013dropout}. This section describes two types of noise injection techniques, such as additive Gaussian noise and dropout noise. 

Let $\mathbf{h}^{(l)}$ denote the $l^{th}$ hidden variables in a neural network, and this layer can be replaced with a noisy version $\tilde{\mathbf{h}}^{(l)}$. We can vary the noise types as the followings. 
\begin{itemize}
\item{
Additive Gaussian noise: $\tilde{\mathbf{h}}^{(l)}=\mathbf{h}^{(l)} + \boldsymbol{\gamma}$, where $\boldsymbol{\gamma} \sim \mathcal{N}(0, \sigma^2\mathbf{I}_{d \times d})$ with the parameter $\sigma^2$ to restrict the degree of noises.
}
\item{
Dropout noise: $\tilde{\mathbf{h}}^{(l)}=\mathbf{h}^{(l)} \odot \boldsymbol{\epsilon} $, where $\odot$ is the elementwise product of two vectors, and the elements of the noise vector are $\boldsymbol{\epsilon}_i \sim Bernoulli(1-p)$ with the parameter $p$. To simply put, this function specifies that $\boldsymbol{\epsilon}_i=0$ with probability $p$ and $\boldsymbol{\epsilon}_i=1$ with probability $(1-p)$.
}
\end{itemize}
Both additive Gaussian noise and dropout noise are generalization techniques to increase the generality of the trained model, but they have different properties. The additive Gaussian noise increases the margin of decision boundaries while the dropout noise affects a model to be sparse \cite{srivastava2014dropout}. These noise layers can be easily included in a deep neural network. For example, there can be a dropout layer between two convolutional layers. Similarly, a layer of additive Gaussian noise can be placed on the input layer. 

\subsection{Self-Ensembling Model}
The recently reported self-ensembling (SE) \cite{laine2016temporal}, or $\Pi$ model, construct a loss function that minimizes the divergence between two outputs from two sampled dropout neural networks with the same input stimulus. Their suggested regularization term can be interpreted as the following:
\begin{equation} \mathcal{L}_{SE}(\mathbf{x}; \boldsymbol{\theta}):= D[f_{\boldsymbol{\theta}}(\mathbf{x}, \boldsymbol{\epsilon}^1), f_{\boldsymbol{\theta}}(\mathbf{x}, \boldsymbol{\epsilon}^2)], \end{equation}
where $\boldsymbol{\epsilon}^1$ and $\boldsymbol{\epsilon}^2$ are randomly sampled dropout noises in a neural network $f_{\boldsymbol{\theta}}$, whose parameters are $\boldsymbol{\theta}$. Also, $D[\mathbf{y}, \mathbf{y}']$ is a non-negative function that represents the distance between two output vectors: $\mathbf{y}$ and $\mathbf{y}'$. For example, $D$ can be the cross entropy function, $D[\mathbf{y}, \mathbf{y}']=-\sum_{i} \mathbf{y}_i  \log \mathbf{y}'_i$, where $\mathbf{y}$ and $\mathbf{y}'$ are the vectors whose $i^{th}$ elements represent the probability of the $i^{th}$ class. The divergence could be calculated between two outputs of two different structures, which turn this regularization to be semi-supervised. $\Pi$ model is based on the principle of $\Gamma$ model, which is the ladder network by Rasmus et al. (\citeyear{rasmus2015semi}). Our proposed method, adversarial dropout, can be seen as a special case of $\Pi$ model when one dropout neural network is adversarially sampled.

\subsection{Adversarial Training}
Adversarial dropout follows the training mechanism of adversarial training, so we briefly introduce a generalized formulation of the adversarial training. The basic concept of adversarial training (AT) is an incorporation of adversarial examples on the training process. Additional loss function by including adversarial examples \cite{szegedy2013intriguing,goodfellow2014explaining,miyato2015distributional} can be defined as a generalized form:
\begin{gather} \mathcal{L}_{AT}(\mathbf{x}, y; \boldsymbol{\theta},  \delta):= D[g(\mathbf{x}, y, \boldsymbol{\theta}), f_{\boldsymbol{\theta}}(\mathbf{x}+\boldsymbol{\gamma}^{adv})] \\ \text{where} \: \boldsymbol{\gamma}^{adv}:=argmax_{\boldsymbol{\gamma};\| \boldsymbol{\gamma}\|_{\infty} \leq \delta} D[g(\mathbf{x}, y, \boldsymbol{\theta}), f_{\boldsymbol{\theta}}(\mathbf{x}+\boldsymbol{\gamma})]. \nonumber \end{gather}
Here, $\boldsymbol{\theta}$ is a set of model parameters,  $\delta$ is a hyperparameter controlling the intensity of the adversarial perturbation $\boldsymbol{\gamma}^{adv}$. The function $f_{\boldsymbol{\theta}}(\mathbf{x})$ is an output distribution of a neural network to be learned. Adversarial training can be diversified by differentiating the definition of $g(\mathbf{x}, y, \boldsymbol{\theta})$, as the following. 
\begin{itemize}
\item{
\emph{Adversarial training (AT)} \cite{goodfellow2014explaining,kurakin2016adversarial} defines $g(\mathbf{x}, y, \boldsymbol{\theta})$ as $g(y)$ ignoring $\mathbf{x}$ and $\boldsymbol{\theta}$, so $g(y)$ is an one-hot encoding vector of $y$. 
}
\item{
\emph{Virtual adversarial training (VAT)} \cite{miyato2015distributional,miyato2016virtual} defines $g(\mathbf{x}, y, \boldsymbol{\theta})$ as $f_{\hat{\boldsymbol{\theta}}}(\mathbf{x})$ where $\hat{\boldsymbol{\theta}}$ is the current estimated parameter. This training method does not use any information from $y$ in the adversarial part of the loss function. This enables the adversarial part to be used as a regularization term for the semi-supervised learning.  
}
\end{itemize}

\section{Method}
This section presents our adversarial dropout that combines the ideas of adversarial training and dropout. First, we formally define the adversarial dropout. Second, we propose a training algorithm to find the instantiations of adversarial dropouts with a fast approximation method. 
\subsection{General Expression of Adversarial Dropout}
Now, we propose the adversarial dropout (AdD), which could be an adversarial training method that determines the dropout condition to be sensitive on the model's label assignment. We use $f_{\boldsymbol{\theta}}(\mathbf{x}, \boldsymbol{\epsilon})$ as an output distribution of a neural  network with a dropout mask. The below is the description of the additional loss function by incorporating adversarial dropout.
\begin{gather}
\label{eq_general}
\mathcal{L}_{AdD}(\mathbf{x}, y, \boldsymbol{\epsilon}^s; \boldsymbol{\theta}, \delta):= D[g(\mathbf{x}, y, \boldsymbol{\theta}), f_{\boldsymbol{\theta}}(\mathbf{x}, \boldsymbol{\epsilon}^{adv})] \\
\text{where} \: \boldsymbol{\epsilon}^{adv}:=argmax_{{\boldsymbol{\epsilon}} ; \| \boldsymbol{\epsilon}^s - \boldsymbol{\epsilon} \|_2 \leq \delta H} D[g(\mathbf{x}, y, \boldsymbol{\theta}), f_{\boldsymbol{\theta}}(\mathbf{x}, \boldsymbol{\epsilon})]. \nonumber
\end{gather}
Here, $D[\cdot,\cdot]$ indicates a divergence function; $g(\mathbf{x}, y, \boldsymbol{\theta})$ represents an adversarial target function that can be diversified by its definition; $\boldsymbol{\epsilon}^{adv}$ is an adversarial dropout mask under the function $f_{\boldsymbol{\theta}}$ when $\boldsymbol{\theta}$ is a set of model parameters; $\boldsymbol{\epsilon}^s$ is a sampled random dropout mask instance; $\delta$ is a hyperparameter controlling the intensity of the noise; and $H$ is the dropout layer dimension. 

We introduce the boundary condition, $\| \boldsymbol{\epsilon}^s - \boldsymbol{\epsilon} \|_2 \leq \delta H$, which indicates a restriction of the number of the difference between two dropout conditions. An adversarial dropout mask should be infinitesimally different from the random dropout mask. Without this constraint, the network with adversarial dropout may become a neural network layer without connections. By restricting the adversarial dropout with the random dropout, we prevent finding such irrational layer, which does not support the back propagation. We found that the Euclidean distance between two $\boldsymbol{\epsilon}$ vectors can be calculated by using the graph edit distance or the Jaccard distance. In the supplementary material, we proved that the graph edit distance and the Jaccard distance can be abstracted as Euclidean distances between two $\boldsymbol{\epsilon}$ vectors. 

In the general form of adversarial training, the key point is the existence of the linear perturbation $\boldsymbol{\gamma}^{adv}$. We can interpret the input with the adversarial perturbation as this adversarial noise input $\tilde{\mathbf{x}}^{adv}=\mathbf{x}+\boldsymbol{\gamma}^{adv}$. From this perspective, the authors of adversarial training limited the adversarial direction only on the space of the additive Gaussian noise $\tilde{\mathbf{x}}=\mathbf{x}+\boldsymbol{\gamma}^{0}$, where $\boldsymbol{\gamma}^{0}$ is a sampled Gaussian noise on the input layer.  
In contrast, adversarial dropout can be considered as a noise space generated by masking hidden units, $\tilde{\mathbf{h}}^{adv}=\mathbf{h} \odot \boldsymbol{\epsilon}^{adv}$ where $\mathbf{h}$ is hidden units, and $\boldsymbol{\epsilon}^{adv}$ is an adversarially selected dropout condition. If we assume the adversarial training as the Gaussian additive perturbation on the input, the perturbation is linear in nature, but adversarial dropout could be non-linear perturbation if the adversarial dropout is imposed upon multiple layers. 

\subsubsection{Supervised Adversarial Dropout}
\emph{Supervised Adversarial dropout (SAdD)} defines $g(\mathbf{x}, y, \boldsymbol{\theta})$ as $y$, so $g$ is a one-hot vector of $y$ as the typical neural network. 
The divergence term from Formula \ref{eq_general} can be converted as follows:
\begin{gather} \mathcal{L}_{SAdD}(\mathbf{x}, y, \boldsymbol{\epsilon}^s; \boldsymbol{\theta}, \delta):= D[g(y), f_{\boldsymbol{\theta}}(\mathbf{x}, \boldsymbol{\epsilon}^{adv})] \\ \text{where} \: \boldsymbol{\epsilon}^{adv}:=argmax_{\boldsymbol{\epsilon} ;\| \boldsymbol{\epsilon}^s - \boldsymbol{\epsilon} \|_2 \leq \delta H} D[g(y), f_{\boldsymbol{\theta}}(\mathbf{x}, \boldsymbol{\epsilon})]. \nonumber \end{gather}
In this case, the divergence term can be seen as the pure loss function for a supervised learning with a dropout regularization.  However, $\boldsymbol{\epsilon}^{adv}$ is selected to maximize the divergence between the true information and the output from the dropout network, so $\boldsymbol{\epsilon}^{adv}$ eventually becomes the mask on the most contributing features. This adversarial mask provides the learning opportunity on neurons, so called \emph{dead filter}, that was considered to be less informative. 

\subsubsection{Virtual Adversarial Dropout}
\emph{Virtual adversarial dropout (VAdD)} defines $g(\mathbf{x}, y, \boldsymbol{\theta})=f_{\boldsymbol{\theta}}(\mathbf{x}, \boldsymbol{\epsilon}^{s})$. This uses the loss function as a regularization term for semi-supervised learning. 
The divergence term in Formula \ref{eq_general} can be represented as bellow:
 \begin{gather} \mathcal{L}_{VAdD}(\mathbf{x}, y, \boldsymbol{\epsilon}^s; \boldsymbol{\theta}, \delta):= D[f_{\theta}(\mathbf{x}, \boldsymbol{\epsilon}^{s}), f_{\boldsymbol{\theta}}(\mathbf{x}, \boldsymbol{\epsilon}^{adv})] \\ \text{where} \: \boldsymbol{\epsilon}^{adv}:=argmax_{\boldsymbol{\epsilon} ;\| \boldsymbol{\epsilon}^s - \boldsymbol{\epsilon} \|_2 \leq \delta H} D[f_{\boldsymbol{\theta}}(\mathbf{x}, \boldsymbol{\epsilon}^{s}), f_{\boldsymbol{\theta}}(\mathbf{x}, \boldsymbol{\epsilon})]. \nonumber \end{gather}
VAdD is a special case of a self-ensembling model with two dropouts. They are 1) a dropout, $\boldsymbol{\epsilon}^s$, sampled from a random distribution with a hyperparameter and 2) a dropout, $\boldsymbol{\epsilon}^{adv}$, composed to maximize the divergence function of the learner, which is the concept of the noise injection from the virtual adversarial training. The two dropouts create a regularization as the virtual adversarial training, and the inference procedure optimizes the parameters to reduce the divergence between the random dropout and the adversarial dropout. This optimization triggers the self-ensemble learning in \cite{laine2016temporal}. However, the adversarial dropout is different from the previous self-ensembling because one dropout is induced by the adversarial setting, not by a random sampling. 

\subsubsection{Learning with Adversarial Dropout}
The full objective function for the learning with the adversarial dropout is given by
\begin{equation} l(y, f_{\boldsymbol{\theta}}(\mathbf{x}, \boldsymbol{\epsilon}^{s})) + \lambda \mathcal{L}_{AdD}(\mathbf{x}, y, \boldsymbol{\epsilon}^{s}; \boldsymbol{\theta}, \delta) \end{equation}
where  $l(y, f_{\boldsymbol{\theta}}(\mathbf{x}, \boldsymbol{\epsilon}^{s}))$ is the negative log-likelihood for $y$ given $x$ under the sampled dropout instance $\boldsymbol{\epsilon}^{s}$. There are two scalar-scale hyper-parameters: (1)  a trade-off parameter, $\lambda$, for controlling the impact of the proposed regularization term and (2)  the constraints, $\delta$, specifying the intensity of adversarial dropout. 

\subsubsection{Combining Adversarial Dropout and Adversarial Training}
Additionally, it should be noted that the adversarial training and the adversarial dropout are not exclusive training methods. A neural network can be trained by imposing the input perturbation with the Gaussian additive noise, and by enabling the adversarially chosen dropouts, simultaneously. Formula \ref{eq_ad_at} specifies the loss function of simultaneously utilizing the adversarial dropout and the adversarial training. 
\begin{equation} \label{eq_ad_at} l(y, f_{\boldsymbol{\theta}}(\mathbf{x}, \boldsymbol{\epsilon}^{s})) + \lambda_{1} \mathcal{L}_{AdD}(\mathbf{x}, y, \boldsymbol{\epsilon}^s) + \lambda_{2} \mathcal{L}_{AT}(\mathbf{x}, y)\end{equation}
where $\lambda_{1}$ and $\lambda_{2}$ are trade-off parameters controlling the impact of the regularization terms. 

\subsection{Fast Approximation Method for Finding Adversarial Dropout Condition}
Once the adversarial dropout, $\boldsymbol{\epsilon}^{adv}$, is identified, the evaluation of $\mathcal{L}_{AdD}$ simply becomes the computation of the loss and the divergence functions. However, the inference on $\boldsymbol{\epsilon}^{adv}$ is difficult because of three reasons. First, we cannot obtain a closed-form solution on the exact adversarial noise value, $\boldsymbol{\epsilon}^{adv}$. Second, the feasible space for $\boldsymbol{\epsilon}^{adv}$ is restricted under $\| \boldsymbol{\epsilon}^s - \boldsymbol{\epsilon}^{adv} \|_2 \leq \delta H$, which becomes a constraint in the optimization.  Third, $\boldsymbol{\epsilon}^{adv}$ is a binary-valued vector rather than a continuous-valued vector because $\boldsymbol{\epsilon}^{adv}$ indicates the activation of neurons. This discrete nature requires an optimization technique like \emph{integer programming}.

To mitigate this difficulty, we approximated the objective function, $\mathcal{L}_{AdD}$, with the first order of the Taylor expansion by relaxing the domain space of $\boldsymbol{\epsilon}^{adv}$. This Taylor expansion of the objective function was used in the earlier works of adversarial training \cite{goodfellow2014explaining,miyato2015distributional}. After the approximation, we found an adversarial dropout condition by solving an integer programming problem. 

To define a neural network with a dropout layer, we separate the output function into two neural sub-networks, $f_{\boldsymbol{\theta}}(\mathbf{x}, \boldsymbol{\epsilon})=f^{upper}_{\boldsymbol{\theta}_1}(\mathbf{h}(\mathbf{x})\odot\boldsymbol{\epsilon})$, where $f^{upper}_{\boldsymbol{\theta}_1}$ is the upper part neural network of the dropout layer and $\mathbf{h}(\mathbf{x})=f^{under}_{\boldsymbol{\theta}_2}(\mathbf{x})$ is the under part neural network. Our objective is optimizing an adversarial dropout noise $\boldsymbol{\epsilon}^{adv}$ by maximizing the following divergence function under the constraint $\| \boldsymbol{\epsilon}^s - \boldsymbol{\epsilon}^{adv} \|_2 \leq \delta H$: 
\begin{equation}
D(\mathbf{x}, \boldsymbol{\epsilon}; \boldsymbol{\theta},\boldsymbol{\epsilon}^{s}) = D[g(\mathbf{x}, y, \boldsymbol{\theta}, \boldsymbol{\epsilon}^s), f^{upper}_{\boldsymbol{\theta}_1}(\mathbf{h}(\mathbf{x})\odot\boldsymbol{\epsilon}))]
\end{equation}
where $\boldsymbol{\epsilon}^{s}$ is a sampled dropout mask, and $\boldsymbol{\theta}$ is a parameter of the neural network model. We approximate the above divergence function by deriving the first order of the Taylor expansion by relaxing the domain space of $\boldsymbol{\epsilon}$ from the multiple binary spaces, $\{0,1\}^{H}$, to the real value spaces, $[0,1]^H$. This conversion is a common step in the integer programming research as \cite{hemmecke2010nonlinear}:
\begin{align}
D(\mathbf{x}, \boldsymbol{\epsilon}; \boldsymbol{\theta},\boldsymbol{\epsilon}^{s}) \approx  D(\mathbf{x}, \boldsymbol{\epsilon}^0; \boldsymbol{\theta},\boldsymbol{\epsilon}^{s}) + (\boldsymbol{\epsilon}-\boldsymbol{\epsilon}^{0})^T \mathbf{J}(\mathbf{x}, \boldsymbol{\epsilon}^{0})   
\end{align}
where $\mathbf{J}(\mathbf{x}, \boldsymbol{\epsilon}^{0})$ is the Jacobian vector given by $\mathbf{J}(\mathbf{x}, \boldsymbol{\epsilon}^{0}):=\boldsymbol{\bigtriangledown}_{\boldsymbol{\epsilon}}D(\mathbf{x}, \boldsymbol{\epsilon}; \boldsymbol{\theta},\boldsymbol{\epsilon}^{s})|_{\boldsymbol{\epsilon}=\boldsymbol{\epsilon}^{0}}$ when $\boldsymbol{\epsilon}^{0}=1$ indicates no noise injection. The above Taylor expansion provides a linearized optimization objective function by controlling $\epsilon$. Therefore, we reorganized the Taylor expansion with respect to $\epsilon$ as the below:
\begin{equation}
\label{Jacobian_def}
D(\mathbf{x}, \boldsymbol{\epsilon}; \boldsymbol{\theta},\boldsymbol{\epsilon}^{s}) \propto \sum_{i} \boldsymbol{\epsilon}_i \mathbf{J}_i(\mathbf{x},\boldsymbol{\epsilon}^0)
\end{equation}
where $\mathbf{J}_i(\mathbf{x},\boldsymbol{\epsilon}^0)$ is the $i^{th}$ element of $\mathbf{J}(\mathbf{x},\boldsymbol{\epsilon}^0)$.
Since we cannot proceed further with the given formula, we introduce an alternative Jaccobian formula that further specifies the dropout mechanism by $\odot$ and $\mathbf{h(x)}$ as the below.
\begin{gather}
J(\mathbf{x},\boldsymbol{\epsilon}^0) \approx \mathbf{h}(\mathbf{x}) \odot \boldsymbol{\bigtriangledown}_{\mathbf{h}(\mathbf{x})} D(\mathbf{x}, \boldsymbol{\epsilon}^{0}; \boldsymbol{\theta}, \boldsymbol{\epsilon}^{s})
\end{gather}
where $\mathbf{h}(\mathbf{x})$ is the output vector of the under part neural network of the adversarial dropout. 

\begin{algorithm}[t]
\caption{Finding Adversarial Dropout Condition}\label{fast_algo}
\SetKwInOut{Input}{Input}
\SetKwInOut{Output}{Output}

\Input{$\boldsymbol{\epsilon}^{s}$ is current sampled dropout mask}
\Input{$\delta$ is a hyper-parameter for the boundary}
\Input{$\mathbf{J}$ is the Jacobian vector}
\Input{$H$ is the layer dimension.}
\Output{$\boldsymbol{\epsilon}_{adv}$}
\Begin{
	
	$\boldsymbol{z} \longleftarrow |\mathbf{J}|$   // absolute values of the Jacobian\\ 
	$\boldsymbol{i} \longleftarrow $ Arg Sort $\boldsymbol{z}$ as $z_{i_1} \leq ... \leq z_{i_H}$  \\
	$\boldsymbol{\epsilon}^{adv} \longleftarrow \boldsymbol{\epsilon}^{s}$ \\
	$d \longleftarrow 1$ \\
	\While{$\| \boldsymbol{\epsilon}^s - \boldsymbol{\epsilon}^{adv} \|_2 \leq \delta H$ and $d \leq H$}{
		\uIf{$\epsilon^{adv}_{i_d}=0$ and $\mathbf{J}_{i_d}>0$}{
			$\epsilon^{adv}_{i_d} \longleftarrow 1$
		}
		\ElseIf{$\epsilon^{adv}_{i_d}=1$ and $\mathbf{J}_{i_d}<0$}{
			$\epsilon^{adv}_{i_d} \longleftarrow 0$
		}
		$d \longleftarrow d + 1$
	}
}
\end{algorithm}

The control variable, $\boldsymbol{\epsilon}$, is a binary vector whose elements are either one or zero. Under this approximate divergence, finding a maximal point of $\boldsymbol{\epsilon}$ can be viewed as the 0/1 knapsack problem \cite{kellerer2004introduction}, which is one of the most popular integer programming problems.  

To find $\boldsymbol{\epsilon}^{adv}$ with the constraint, we propose Algorithm \ref{fast_algo} based on the dynamic programming for the 0/1 knapsack problem. In the algorithm, $\boldsymbol{\epsilon}^{adv}$ is initialized with $\boldsymbol{\epsilon}^s$, and $\boldsymbol{\epsilon}^{adv}$ changes its value by the order of the degree increasing the objective divergence until $\| \boldsymbol{\epsilon}^s - \boldsymbol{\epsilon}^{adv} \|_2 \leq \delta H$; or there is no increment in the divergence. After using the algorithm, we obtain $\boldsymbol{\epsilon}^{adv}$ that maximizes the divergence with the constraint, and we evaluate the loss function $\mathcal{L}_{AdD}$.

We should notice that the complex vector of the Taylor expansion is not $\boldsymbol{\epsilon}^{s}$, but $\boldsymbol{\epsilon}^{0}$. In the case of virtual adversarial dropout, whose divergence is formed as $D[f_{\boldsymbol{\theta}}(\mathbf{x}, \boldsymbol{\epsilon}^s), f_{\boldsymbol{\theta}}(\mathbf{x}, \boldsymbol{\epsilon})]$, $\boldsymbol{\epsilon}^{s}$ is the minimal point leading the gradient to be zero because of the identical distribution between the random and the optimized dropouts. This zero gradient affects the approximation of the divergence term as zero. To avoid the zero gradients, we set the complex vector of the Taylor expansion as $\boldsymbol{\epsilon}_{0}$. 

This zero gradient situation does not occur when the model function, $f_{\boldsymbol{\theta}}$, contains additional stochastic layers because $f_{\boldsymbol{\theta}}(\mathbf{x}, \boldsymbol{\epsilon}^s, \boldsymbol{\rho}^1) \neq f_{\boldsymbol{\theta}}(\mathbf{x}, \boldsymbol{\epsilon}^s, \boldsymbol{\rho}^2)$ when $\boldsymbol{\rho}^1$ and $\boldsymbol{\rho}^2$ are independently sampled noises from another stochastic layers.  

\begin{table}[t]
  \caption{Test performance with 1,000 labeled (semi-supervised) and 60,000 labeled (supervised) examples on MNIST. Each setting is repeated for eight times. }
  \label{cnn_mnist_sup}
  \centering
  \begin{tabular}{lclcl}
    \hline
    										& \multicolumn{2}{c}{Error rate ($\%$) with $\#$ labels }                   \\
												\cline{2-3}
    Method     								& 1,000 				& All (60,000)   \\
    \hline
    Plain (only dropout) 						&  2.99 $\pm$ 0.23 		&  0.53 $\pm$ 0.03     \\
    AT	     						&  		-	 		& 0.51 $\pm$ 0.03      \\
    VAT     				&  1.35 $\pm$ 0.14		& 0.50 $\pm$ 0.01     \\
    $\Pi$ model    							&  1.00 $\pm$ 0.08 		& 0.50 $\pm$ 0.02      \\
    \hline 
    SAdD    									&  		- 			& \textbf{0.46} $\pm$ \textbf{0.01}      \\
    VAdD (KL)    								&  \textbf{0.99} $\pm$ \textbf{0.07} 		& 0.47 $\pm$ 0.01      \\
    VAdD (QE)    								&  \textbf{0.99} $\pm$ \textbf{0.09} 		& \textbf{0.46} $\pm$ \textbf{0.02}      \\
    \hline 
  \end{tabular}
\end{table}

\section{Experiments}

This section evaluates the empirical performance of adversarial dropout for supervised and semi-supervised classification tasks on three benchmark datasets, MNIST, SVHN, and CIFAR-10. In every presented task, we compared adversarial dropout, $\Pi$ model, and adversarial training. We also performed additional experiments to analyze the sparsity of adversarial dropout.    

\subsection{Supervised and Semi-supervised Learning on MNIST task}
In the first set of experiments, we benchmark our method on the MNIST dataset \cite{lecun1998gradient}, which consists of 70,000 handwritten digit images of size $28\times28$ where 60,000 images are used for training and the rest for testing. 

Our basic structure is a convolutional neural network (CNN) containing three convolutional layers, which filters are 32, 64, and 128, respectively, and three max-pooling layers sized by $2 \times 2$. The adversarial dropout applied only on the final hidden layer. The structure detail and the hyper-parameters are described in Appendix B.1. 

We conducted both supervised and semi-supervised learnings to compare the performances from the standard dropout, $\Pi$ model, and adversarial training models utilizing linear perturbations on the input space. The supervised learning used 60,000 instances for training with full labels. The semi-supervised learning used 1,000 randomly selected instances with their labels and 59,000 instances with only their input images. Table 1 shows the test error rates including the baseline models. Over all experiment settings, SAdD and VAdD further reduce the error rate from $\Pi$ model, which had the best performance among the baseline models. In the table, KL and QE indicate Kullback-Leibler divergence and quadratic error, respectively, to specify the divergence function, $D[\mathbf{y}, \hat{\mathbf{y}}]$. 

%
%
\subsection{Supervised and Semi-supervised Learning on SVHN and CIFAR-10}

\begin{table*}[t]
  \caption{Test performances of semi-supervised and supervised learning on SVHN and CIFAR-10. Each setting is repeated for five times. KL and QE indicate Kullback-Leibler divergence and quadratic error, respectively, to specify the divergence function, $D[\mathbf{y}, \hat{\mathbf{y}}]$}
  \label{cnn_mnist_sup}
  \centering
  \begin{tabular}{lccccc}
    \hline
    										& \multicolumn{2}{c}{SVHN with $\#$ labels }  & & \multicolumn{2}{c}{CIFAR-10 with $\#$ labels }                 \\
												\cline{2-3} \cline{5-6}
    Method     										& 1,000 	& 73,257 (All)   			&& 4,000 	& 50,000 (All)   \\
    \hline
    $\Pi$ model \cite{laine2016temporal}     				&  4.82     & 2.54					&&  12.36     & 5.56  \\ 
    Tem. ensembling \cite{laine2016temporal}     			&  4.42     & 2.74  				&&  12.16     & 5.60  \\ 
    Sajjadi et al. \cite{sajjadi2016regularization} 			&- &- 						&&  11.29     & - \\ 
    VAT \cite{miyato2017virtual}     					&  3.86     & - 					&&  10.55     & 5.81 \\ 
    \hline
    $\Pi$ model (our implementation)     			&   4.35  $\pm$ 0.04     &   2.53 $\pm$ 0.05	&&   12.62  $\pm$ 0.29     &   5.77 $\pm$ 0.11\\
    VAT (our implementation)					&   \textbf{3.74} $\pm$ \textbf{0.09}    &   2.69 $\pm$ 0.04  	&&   11.96 $\pm$ 0.10    &   5.65 $\pm$ 0.17  \\
    SAdD 							&   - 				  &   2.46 $\pm$ 0.05 	&&   - 				  &   5.46 $\pm$ 0.16 \\
    VAdD (KL)						&   4.16 $\pm$ 0.08   &   \textbf{2.31} $\pm$ \textbf{0.01} 	&&   11.68 $\pm$ 0.19    &   5.27 $\pm$ 0.10 \\
    VAdD (QE)						&   4.26 $\pm$ 0.14    &   2.37 $\pm$ 0.03 	&&   \textbf{11.32} $\pm$ \textbf{0.11}    &   \textbf{5.24} $\pm$ \textbf{0.12} \\
   \hline
    VAdD (KL) + VAT	&   \textbf{3.55} $\pm$ \textbf{0.05} 	& \textbf{2.23} $\pm$ \textbf{0.03} 	&&   10.07 $\pm$ 0.11    & \textbf{4.40} $\pm$ \textbf{0.12} \\
    VAdD (QE) + VAT	&   \textbf{3.55} $\pm$ \textbf{0.07} & 2.34 $\pm$ 0.05    				&&   \textbf{9.22} $\pm$ \textbf{0.10} & 4.73 $\pm$ 0.04    \\
    \hline 
  \end{tabular}
\end{table*}

We experimented the performances of the supervised and the semi-supervised tasks on the SVHN \cite{netzer2011reading} and the CIFAR-10 \cite{krizhevsky2009learning} datasets consisting of $32\times32$ color images in ten classes. For these experiments, we used the large-CNN  \cite{laine2016temporal,miyato2017virtual}.  The details of the structure and the settings are described in Appendix B.2.

Table \ref{cnn_mnist_sup} shows the reported performances of the close family of CNN-based classifiers for the supervised and semi-supervised learning. We did not consider the recently advanced architectures, such as ResNet \cite{he2016identity} and DenseNet \cite{huang2016densely}, because we intend to compare the performance increment by the dropout and other training techniques. 

In supervised learning tasks using all labeled train data, adversarial dropout models achieved the top performance compared to the results from the baseline models, such as $\Pi$ model and VAT, on both datasets. When applying adversarial dropout and adversarial training together, there were further improvements in the performances.

Additionally, we conducted experiments on the semi-supervised learning with randomly selected labeled data and unlabeled images. In SVHN, 1,000 labeled and 72,257 unlabeled data were used for training. In CIFAR-10, 4,000 labeled and 46,000 unlabeled data were used. Table \ref{cnn_mnist_sup} lists the performance of the semi-supervised learning models, and our implementations with both VAdD and VAT achieved the top performance compared to the results from \cite{sajjadi2016regularization}. 

Our experiments demonstrate that VAT and VAdD are complementary. When applying VAT and VAdD together by simply adding their divergence terms on the loss function, see Formula \ref{eq_ad_at}, we achieved the state-of-the-art performances on the semi-supervised learning on both datasets; 3.55\% of test error rates on SVHN, and 10.04\% and  9.22\% of test error rates on CIFAR-10. Additionally, VAdD alone achieved a better performance than the self-ensemble model ($\Pi$ model). This indicates that considering an adversarial perturbation on dropout layers enhances the self-ensemble effect. 

\subsection{Effect on Features and Sparsity from Adversarial Dropout}
Dropout prevents the co-adaptation between the units in a neural network, and the dropout decreases the dependency between hidden units \cite{srivastava2014dropout}. To compare the adversarial dropout and the standard dropout, we analyzed the co-adaptations by visualizing features of autoencoders on the MNIST dataset. The autoencoder consists with one hidden layer, whose dimension is 256, with the ReLU activation. When we trained the autoencoder, we set the dropout with $p = 0.5$, and we calculated the reconstruction error between the input data and the output layer as a loss function to update the weight values of the autoencoder with the standard dropout. On the other hand, the adversarial dropout error is also considered when we update the weight values of the autoencoder with the parameters, $\lambda$ = 0.2, and $\delta$ = 0.3. The trained autoencoders showed similar reconstruction errors on the test dataset.

\begin{figure}[h]
 \centering
	\includegraphics[width=0.23 \textwidth]{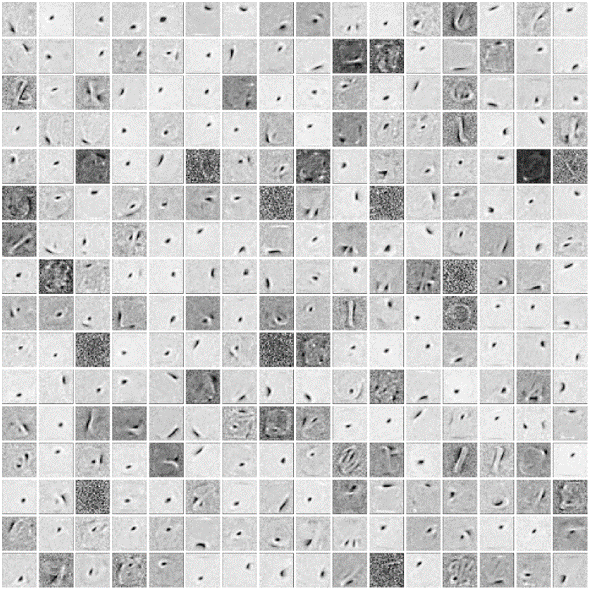}
	\includegraphics[width=0.23 \textwidth]{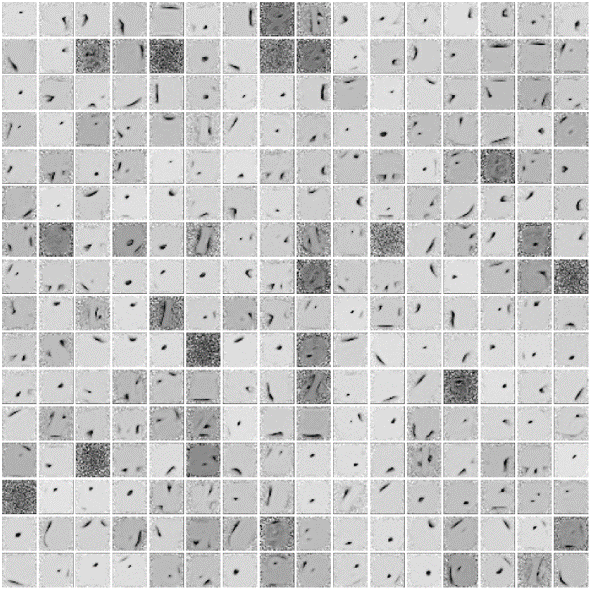}
  \caption{Features of one hidden layer autoencoders trained on MNIST; a standard dropout (left) and an adversarial dropout (right).}
  \label{features}
\end{figure}
Figure \ref{features} shows the visualized features from the autoencoders. There are two differences identified from the visualization; 1) adversarial dropout prevents that the learned weight matrix contains black boxes, or \emph{dead filters}, which may be all zero for many different inputs and 2) adversarial dropout tends to standardize other features, except for localized features viewed as black dots, while the standard dropout tends to ignore the neighborhoods of the localized features. These show that adversarial dropout standardizes the other features while preserving the characteristics of localized features from the standard dropout . These could be the main reason for the better generalization performance.

\begin{figure}[t!]
  \centering
     \includegraphics[width=0.45\textwidth]{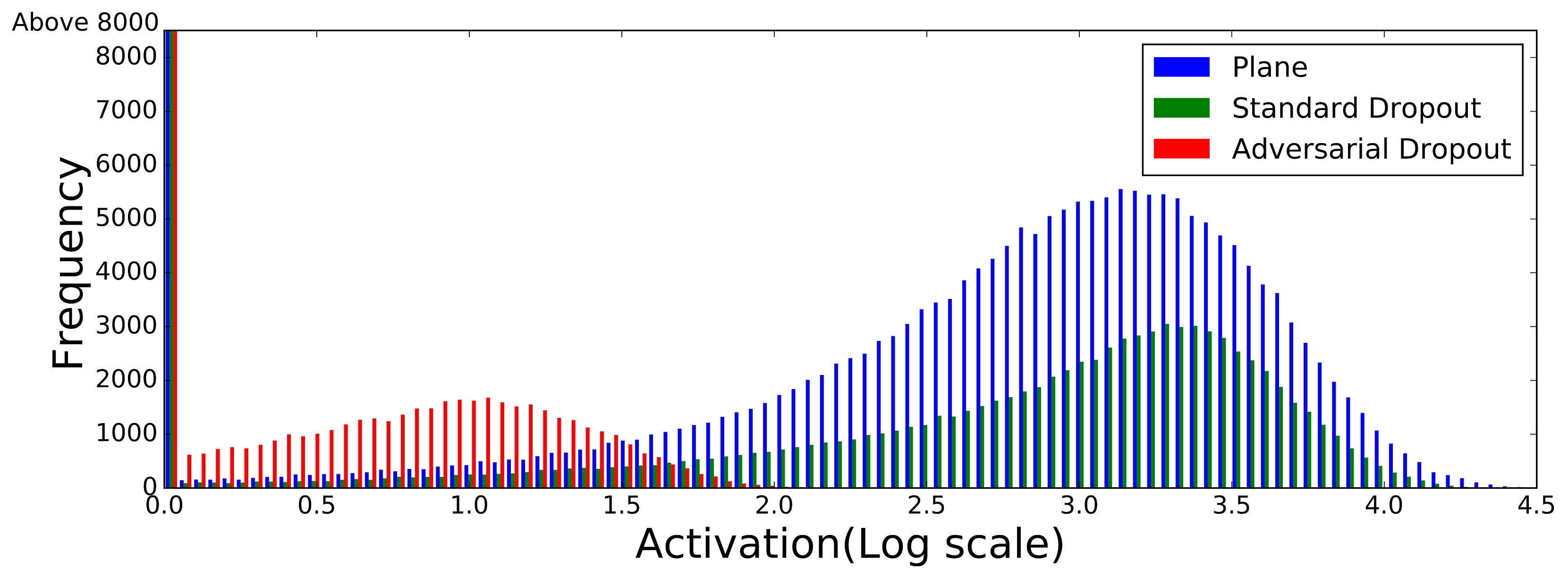}
     \includegraphics[width=0.45\textwidth]{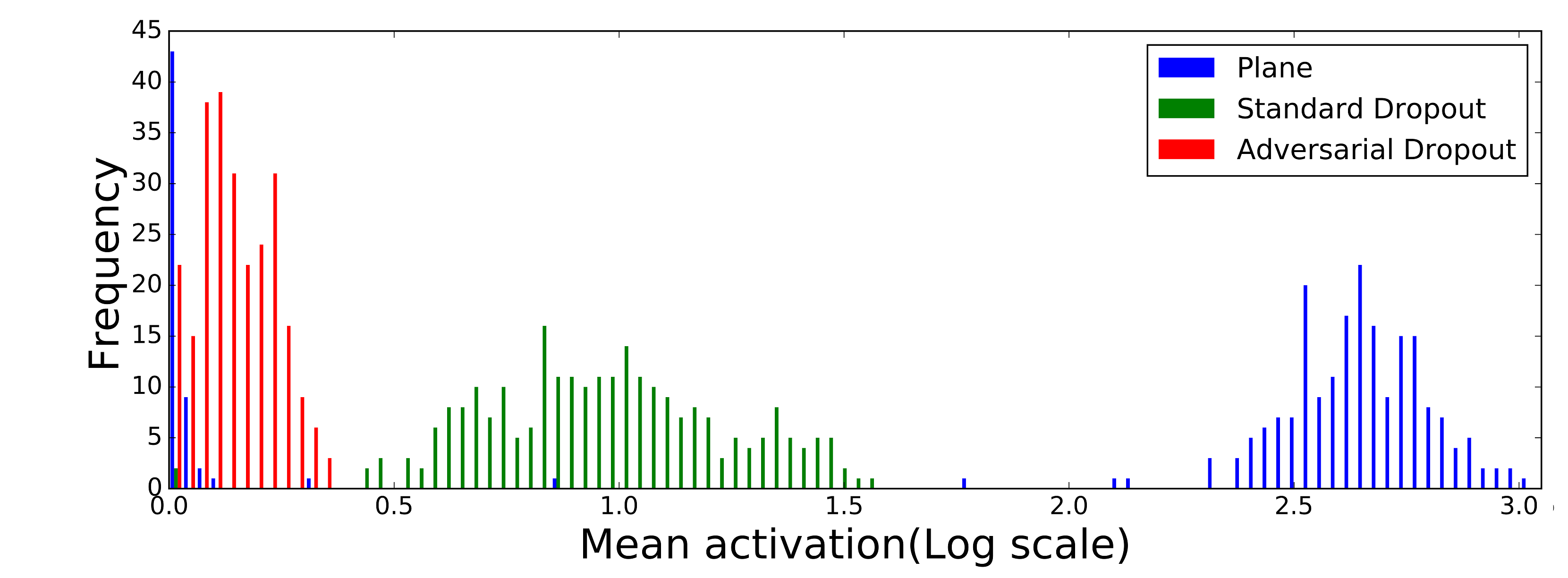}
  \caption{Histograms of  the activation values and the mean activation values from a hidden layer of autoencoders in 1,000 MNIST test images. All values are converted by the log scale for the comparison.} 
  \label{histogram2} 
\end{figure}

The important side-effect of the standard dropout is the sparse activations of the hidden units \cite{hinton2012improving}. To analyze the sparse activations by adversarial dropout, we compared the activation values of the auto-encoder models with no-dropout, dropout, and adversarial dropout on the MNIST test dataset. A sparse model should only have a few highly activated units, and the average activation of any unit across data instances should be low \cite{hinton2012improving}. Figure \ref{histogram2} plot the distribution of the activation values and their means across the test dataset. We found that the adversarial dropout has fewer highly activated units compared to others. Moreover, the mean activation values of the adversarial dropout were the lowest. These indicate that adversarial dropout improves the sparsity of the model than the standard dropout does.

\section{Disucssion}

The previous studies proved that the adversarial noise injections were an effective regularizer \cite{goodfellow2014explaining}. In order to investigate the different properties of adversarial dropout, we explore a very simple case of applying adversarial training and adversarial dropout to the linear regression. 

\subsection{Linear Regression with Adversarial Training}  
Let $\mathbf{x}_i \in \mathbb{R}^{D}$ be a data point and $y_i \in \mathbb{R}$ be a target where $i=\{1,...,N\}$. The objective of the linear regression is finding $\mathbf{w} \in \mathbb{R}^{D}$ that minimizes $l(\mathbf{w})=\sum_i \| y_i - \mathbf{x}_i^T\mathbf{w} \|^{2}$. 

To express adversarial examples, we denote $\tilde{\mathbf{x}}_i = \mathbf{x}_i + \mathbf{r}_i^{adv}$ as the adversarial example of $\mathbf{x}_i$ where $\mathbf{r}_i^{adv}=\delta sign( \boldsymbol{\bigtriangledown}_{\mathbf{x}_i} l(\mathbf{w}))$ utilizing the fast gradient sign method (FGSM) \cite{goodfellow2014explaining}, $\delta$ is a control parameter representing the degree of adversarial noises. With the adversarial examples, the objective function of the adversarial training can be viewed as follows:
\begin{gather}
l_{AT}(\mathbf{w})=\sum_i \| y_i - (\mathbf{x}_i+\mathbf{r}^{adv}_i)^T\mathbf{w} \|^{2}
\end{gather}
The above equation is translated into the below formula by isolating the terms with $\mathbf{r}_i^{adv}$ as the additive noise.
\begin{gather}
l(\mathbf{w}) + \sum_{ij} |\delta \bigtriangledown_{x_{ij}} l(\mathbf{w})| + \delta^2 \mathbf{w}^T \Gamma_{AT} \mathbf{w}
\end{gather}
where $\Gamma_{AT}= \sum_i sign( \boldsymbol{\bigtriangledown}_{\mathbf{x}_i} l(\mathbf{w}))^T sign( \boldsymbol{\bigtriangledown}_{\mathbf{x}_i} l(\mathbf{w}) )$. The second term shows the $L_1$ regularization by multiplying the degree of the adversarial noise, $\delta$, at each data point. Additionally, the third term indicates the $L_2$ regularization with $\Gamma_{AT}$, which form the scales of $\mathbf{w}$ by the gradient direction differences over all data points. The penalty terms are closely related with the hyper-parameter $\delta$. When $\delta$ approaches to zero, the regularization term disappears because the inputs become adversarial examples, not anymore. For a large $\delta$, the regularization constant grows larger than the original loss function, and the learning becomes infeasible. The previous studies proved that the adversarial objective function based on the FGSM is an effective regularizer. This paper investigated that training a linear regression with adversarial examples provides two regularization terms of the above equation. 

\subsection{Linear Regression with Adversarial Dropout}  
Now, we turn to the case of applying adversarial dropout to a linear regression. To represent the adversarial dropout, we denote $\tilde{\mathbf{x}}_i = \boldsymbol{\epsilon}_i^{adv}\odot\mathbf{x}_i$ as the adversarially dropped input of $\mathbf{x}_i$ where $\boldsymbol{\epsilon}_i^{adv}=argmax_{\boldsymbol{\epsilon}; \|\boldsymbol{\epsilon}_i - 1 \|_{2} \leq k} \| y_i - (\boldsymbol{\epsilon}_i \odot \mathbf{x}_i)^T\mathbf{w} \|^{2}$ with the hyper-parameter, $k$, controlling the degree of the adversarial dropout. For simplification, we used one vector as the sampled dropout, $\boldsymbol{\epsilon}^{s}$, of the adversarial dropout. If we apply Algorithm \ref{fast_algo}, the adversarial dropout can be defined as follows:
\begin{gather}
\epsilon_{ij}^{adv} = \left\{ 
\begin{array}{rcl}
 0 &  & \mbox{if} \: \mathbf{x}_{ij}\boldsymbol{\bigtriangledown}_{\mathbf{x}_{ij}} l(\mathbf{w}) \leq min\{s_{ik}, 0\}  \\
 1 &  & \mbox{otherwise} 
\end{array} \right.
\end{gather}
where $s_{ik}$ is the $k^{th}$ lowest element of $ \mathbf{x}_{i} \odot \boldsymbol{\bigtriangledown}_{\mathbf{x}_{i}} l(\mathbf{w})$. This solution satisfies the constraint, $\|\boldsymbol{\epsilon}_i - \boldsymbol{\epsilon}^{s} \|_{2} \leq k$. With this adversarial dropout condition, the objective function of the adversarial dropout can be defined as the belows:
\begin{gather}
l_{AdD}(\mathbf{w})=\sum_i \| y_i - (\boldsymbol{\epsilon}_i^{adv} \odot \mathbf{x}_i)^T\mathbf{w} \|^{2} 
\end{gather}
When we isolate the terms with $\boldsymbol{\epsilon}^{adv}$, the above equation is translated into the below formula.
\begin{gather}
l(\mathbf{w}) + \sum_{i} \sum_{j \in S_{i}} | x_{ij} \bigtriangledown_{x_{ij}} l(\mathbf{w})| +  \mathbf{w}^T \Gamma_{AdD} \mathbf{w}
\end{gather}
where $S_{i} = \{ j | \epsilon_{ij}^{adv}=0 \}$ and  $\Gamma_{AdD}=\sum_{i} (( 1 - \boldsymbol{\epsilon}_i^{adv}) \odot \mathbf{x}_i)^T (( 1 - \boldsymbol{\epsilon}_i^{adv}) \odot \mathbf{x}_i)$. The second term is the $L_1$ regularization of the $k$ largest loss changes from the features of each data point. The third term is the $L_2$ regularization with $\Gamma_{AdD}$. These two penalty terms are related with the hyper-parameter $k$ controlling the degree of the adversarial dropout, because the $k$ indicates the number of elements of the set $S_i, \forall i $. When $k$ becomes zero, the two penalty terms disappears because there will be no dropout by the constraint on $\boldsymbol{\epsilon}$. 

There are two differences between the adversarial dropout and the adversarial training. First, the regularization terms of the adversarial dropout are dependent on the scale of the features of each data point. In $L_1$ regularization, the gradients of the loss function are re-scaled with the data points. In $L_2$ regularization, the data points affect the scales of the weight costs. In contrast, the penalty terms of adversarial training are dependent on the degree of adversarial noise, $\delta$, which is a static term across the instances because $\delta$ is a single-valued hyper parameter given in the training process. Second, the penalty terms of the adversarial dropout are selectively activated by the degree of the loss changes while the penalty terms of the adversarial training are always activated.  

\section{Conclusion}

The key point of our paper is combining the ideas from the adversarial training and the dropout. The existing methods of the adversarial training control a linear perturbation with additive properties only on the input layer. In contrast, we combined the concept of the perturbation with the dropout properties on hidden layers. Adversarially dropped structure becomes a poor ensemble model for the label assignment even when very few nodes are changed. However, by learning the model with the poor structure, the model prevents over-fitting using a few effective features. The experiments showed that the generalization performances are improved by applying our adversarial dropout. Additionally, our approach achieved the-state-of-the-art performances of 3.55\% on SVHN and 9.22\% on CIFAR-10 by applying VAdD and VAT together for the semi-supervised learning. 

\bibliographystyle{aaai}
\bibliography{adt_reff}

\appendix

\section{Appendix A. Distance between Two Dropout Conditions}
In this section, we describe process of induction for the boundary condition from the constraints $distance(\epsilon,\epsilon^{s})<\delta$. We applied two distance metrics, graph edit distance (GED) and Jaccard distance (JD) and proved that restricting upper bounds of two metrics is same with limiting the Euclidean distance. 
\begin{equation}
GED(\boldsymbol{\epsilon}^1,\boldsymbol{\epsilon}^{2}) \propto JD(\boldsymbol{\epsilon}^1,\boldsymbol{\epsilon}^{2}) \propto ||\boldsymbol{\epsilon}^1 - \boldsymbol{\epsilon}^2||_2
\end{equation}
Following sub-sections show the propositions.

\subsection{A.1. Graph Edit Distance}
When we consider a neural network as a graph, we can apply the graph edit distance \cite{sanfeliu1983distance} to measure relative difference between two dropouted networks, $g_1$ and $g_2$, by dropout masks, $\boldsymbol{\epsilon}^1$ and $\boldsymbol{\epsilon}^2$. The following is the definition of graph edit distance (GED) between two networks.
\begin{equation}
GED(g_1, g_2)=min_{(e_1,...,e_k) \in P(g_1, g_2) } \sum_{i=1}^{k} c(e_i),
\end{equation}
where $P(g_1, g_2)$ denotes the set of edit path transforming $g_1$ into $g_2$, $c(e) \geq 0$ is the cost of each graph edit operation $e$, and $k$ is the number of  the edit operations required to change $g_1$ to $g_2$. For simplification, we only considered edge insertion and deletion operations and their cost are same as $1$. When a hidden node (vertex) is dropped, the cost of the GED is $N_l+N_u$ where $N_l$ is the numbers of lower layer nodes and $N_u$ is the number of upper layer nodes. If we consider a hidden node (vertex) is revival, change of GED is same as $N_l+N_u$. This leads following proposition.

\begin{prop}
Given two networks $g_1$ and $g_2$, generated by two dropout masks $\boldsymbol{\epsilon}^1$ and $\boldsymbol{\epsilon}^2$, and all graph edit costs are same as $c(e)=1$, graph edit distance with two dropout masks can be interpreted as:
\begin{equation}
GED(g_1, g_2)=(N_l+N_u) \|\boldsymbol{\epsilon}^1 - \boldsymbol{\epsilon}^2\|_2.
\end{equation} 
Due to $\boldsymbol{\epsilon}^1$ and $\boldsymbol{\epsilon}^2$ are binary masks, their Euclidean distance can provide the number of different dropped nodes.  
\end{prop}

\subsection{A.2. Jaccard Distance}
When we consider a dropout condition $\boldsymbol{\epsilon}$ as a set of selected hidden nodes, we can apply Jaccard distance to measure difference two dropout masks, $\boldsymbol{\epsilon}^1$ and $\boldsymbol{\epsilon}^2$. The following equation is the definition of Jaccard distance:
\begin{equation}
JD(\boldsymbol{\epsilon}^1, \boldsymbol{\epsilon}^2) = \frac{|\boldsymbol{\epsilon}^1 \cup \boldsymbol{\epsilon}^2| - |\boldsymbol{\epsilon}^1 \cap \boldsymbol{\epsilon}^2|}{|\boldsymbol{\epsilon}^1 \cup \boldsymbol{\epsilon}^2|}.
\end{equation}
Since $\boldsymbol{\epsilon}^1$ and $\boldsymbol{\epsilon}^2$ are binary vectors,  $|\boldsymbol{\epsilon}^1 \cap \boldsymbol{\epsilon}^2|$ can be converted as $\|\boldsymbol{\epsilon}^1 \odot \boldsymbol{\epsilon}^2\|_2$ and $|\boldsymbol{\epsilon}^1 \cup \boldsymbol{\epsilon}^2|$ can be viewed as $\|\boldsymbol{\epsilon}^1 + \boldsymbol{\epsilon}^2 - \boldsymbol{\epsilon}^1 \odot \boldsymbol{\epsilon}^2\|_2$. This leads the following proposition. 

\begin{prop}
Given two dropout masks $\boldsymbol{\epsilon}^1$ and $\boldsymbol{\epsilon}^2$, which are binary vectors, Jaccard distance between them can be defined as:
\begin{equation}
JD(\boldsymbol{\epsilon}^1, \boldsymbol{\epsilon}^2)=\frac{\|\boldsymbol{\epsilon}^1 - \boldsymbol{\epsilon}^2\|_2}{\|\boldsymbol{\epsilon}^1 + \boldsymbol{\epsilon}^2 - \boldsymbol{\epsilon}^1 \odot \boldsymbol{\epsilon}^2\|_2}.
\end{equation}
\end{prop} 


\section{Appendix B. Detailed Experiment Set-up}

This section describes the network architectures and settings for the experimental results in this paper. The tensorflow implementations for reproducing these results can be obtained from \url{https://github.com/sungraepark/Adversarial-Dropout}. 
 
\subsection{B.1. MNIST : Convolutional Neural Networks}

\begin{table}[h]
  \caption{The CNN architecture used on MNIST}
  \label{cnn_cifar_architecture}
  \centering
  \begin{tabular}{ll}
    \hline
    Name     								&  Description \\
    \hline 
    input 				&  28 X 28 image     \\ 
    conv1    			&  32 filters, 1 x 1, pad='same', ReLU      \\ 
    pool1				&  Maxpool 2 x 2 pixels \\
    drop1				&  Dropout, $p=0.5$ \\
    conv2   				&  64 filters, 1 x 1, pad='same', ReLU      \\ 
    pool2				&  Maxpool 2 x 2 pixels \\
    drop2				&  Dropout, $p=0.5$ \\
    conv3   				&  128 filters, 1 x 1, pad='same', ReLU      \\ 
    pool3				&  Maxpool 2 x 2 pixels \\
    adt				&  Adversarial dropout, $p=0.5$, $\delta=0.005$ \\
    dense1				&  Fully connected 2048 $\rightarrow$ 625 \\
    dense2				&  Fully connected 625 $\rightarrow$ 10 \\
    output				&  Softmax \\
    \hline 
  \end{tabular}
\end{table}

The MNIST dataset (LeCun et al., 1998) consists of 70,000 handwritten digit images of size $28\times28$ where 60,000 images are used for training and the rest for testing. The CNN architecture is described in Table 1. All networks were trained using Adam \cite{kingma2014adam} with a learning rate of 0.001 and momentum parameters of $\beta_1=0.9$ and $\beta_2=0.999$. In all implementations, we trained the model for 100 epochs with minibatch size of 128. 

For the constraint of adversarial dropout, we set $\delta=0.005$, which indicates 10 ($2048*0.005$) adversarial changes from the randomly selected dropout mask. In all training, we ramped up the trade-off parameter, $\lambda$, for proposed regularization term, $\mathcal{L}_{AdD}$. During the first 30 epochs, we used a Gaussian ramp-up curve $exp[-5(1-T)^2]$, where $T$ advances linearly from zero to one during the ramp-up period. The maximum values of $\lambda_{max}$ are 1.0 for VAdD (KL) and VAT , and 30.0 for VAdD (QE) and $\Pi$ model.  

\subsection{B.2. SVHN and CIFAR-10 : Supervised and Semi-supervised learning}

\begin{table}[h]
  \caption{The network architecture used on SVHN and CIFAR-10}
  \label{cnn_cifar_architecture}
  \centering
  \begin{tabular}{ll}
    \hline

    Name     								&  Description \\
    \hline 
    input 				&  32 X 32 RGB image     \\ 
    noise    				&  Additive Gaussian noise $\sigma=0.15$      \\ 
    conv1a    			&  128 filters, 3 x 3, pad='same', LReLU($\alpha=0.1$)      \\ 
    conv1b    			&  128 filters, 3 x 3, pad='same', LReLU($\alpha=0.1$)      \\ 
    conv1c    			&  128 filters, 3 x 3, pad='same', LReLU($\alpha=0.1$)      \\ 
    pool1				&  Maxpool 2 x 2 pixels \\
    drop1				&  Dropout, $p=0.5$ \\
    conv2a    			&  256 filters, 3 x 3, pad='same', LReLU($\alpha=0.1$)      \\ 
    conv2b    			&  256 filters, 3 x 3, pad='same', LReLU($\alpha=0.1$)      \\ 
    conv2c    			&  256 filters, 3 x 3, pad='same', LReLU($\alpha=0.1$)      \\ 
    pool2				&  Maxpool 2 x 2 pixels \\
    conv3a    			&  512 filters, 3 x 3, pad='valid', LReLU($\alpha=0.1$)      \\ 
    conv3b    			&  256 filters, 1 x 1, LReLU($\alpha=0.1$)      \\ 
    conv3c    			&  128 filters, 1 x 1, LReLU($\alpha=0.1$)      \\ 
    pool3				&  Global average pool (6 x 6 $\rightarrow$ 1 x 1)pixels \\
    add				&  Adversarial dropout, $p=1.0$, $\delta=0.05$ \\
    dense				&  Fully connected 128 $\rightarrow$ 10 \\
    output				&  Softmax \\
    \hline 
  \end{tabular}
\end{table}

The both datasets, SVHN \cite{netzer2011reading} and CIFAR-10  \cite{krizhevsky2009learning}, consist of $32\times32$ colour images in ten classes. For these experiments, we used a CNN, which used by \cite{laine2016temporal,miyato2017virtual} described in Table 2. In all layers, we applied batch normalization for SVHN and mean-only batch normalization \cite{salimans2016weight} for CIFAR-10 with momentum 0.999. All networks were trained using Adam \cite{kingma2014adam} with the momentum parameters of $\beta_1=0.9$ and $\beta_2=0.999$, and the maximum learning rate 0.003. We ramped up the learning rate during the first 80 epochs using a Gaussian ramp-up curve $exp[-5(1-T)^2]$, where  $T$ advances linearly from zero to one during the ramp-up period. Additionally, we annealed the learning rate to zero and the Adam parameter, $\beta_1$, to 0.5 during the last 50 epochs. The number of total epochs is set as 300. These learning setting are same with \cite{laine2016temporal}. 

For adversarial dropout, we set the maximum value of regularization component weight, $\lambda_{max}$, as 1.0 for VAdD(KL) and 25.0 for VAdD(QE). We also ramped up the weight using the Gaussian ramp-up curve during the first 80 epochs. Additionally, we set $\delta$ as 0.05 and dropout probability $p$ as 1.0, which means dropping 6 units among the full hidden units. We set minibatch size as 100 for supervised learning and 32 labeled and 128 unlabeled data for semi-supervised learning.       

\section{Appendix C. Definition of Notation}
In this section, we describe notations used over this paper.  

\begin{table}[h]
  \caption{The notation used over this paper.}
  \label{notation}
  \centering
  \begin{tabular}{ll}
    \hline

    Notat.     								&  Description \\
    \hline
    $\mathbf{x}$				& An input of a neural network \\
    $y$				& A true label  \\
    $\theta$			& A set of parameters of a neural network \\
    $\boldsymbol{\gamma}$			& A noise vector of additive Gaussian noise layer \\
    $\boldsymbol{\epsilon}$			& A binary noise vector of dropout layer \\
    $\delta$			& A hyperparameter controlling the intensity of  \\
					& the adversarial perturbation \\
    $\lambda$			& A trade-off parameter controlling the impact of \\
					& a regularization term \\
    \hline 
    $D[\mathbf{y}, \mathbf{y}']$			& A nonnegative function that represents the  \\
					& distance between two output vectors: \\
           				& cross entropy(CE), KL divergence(KL), and  \\
					& quadratic error (QE) \\
    $f_{\boldsymbol{\theta}}(\mathbf{x})$		& An output vector of a neural network with \\
					& parameters ($\boldsymbol{\theta}$) and an input ($\mathbf{x}$) \\
    $f_{\boldsymbol{\theta}}(\mathbf{x}, \boldsymbol{\rho})$	& An output vector of a neural network with  \\
					& parameters ($\boldsymbol{\theta}$), an input ($\mathbf{x}$), and \\
					& noise ($\boldsymbol{\rho}$) \\
    $f^{upper}_{\boldsymbol{\theta}_1}$ 	& A upper part of a neural network, $f_{\boldsymbol{\theta}}(\mathbf{x}, \boldsymbol{\epsilon})$, \\
					& of a adversarial dropout layer where  \\
					& $\boldsymbol{\theta}=\{ \boldsymbol{\theta}_1, \boldsymbol{\theta}_2 \}$  \\
    $f^{under}_{\boldsymbol{\theta}_2}$   &A under part of a neural network, $f_{\boldsymbol{\theta}}(\mathbf{x}, \boldsymbol{\epsilon})$, \\
					&of a adversarial dropout layer  \\
    \hline 
  \end{tabular}
\end{table}

\section{Appendix D. Performance Comparison with Other Models}

\subsection{D.1. CIFAR-10 : Supervised classification results with additional baselines}

We compared the reported performances of the additional close family of CNN-based classifier for the supervised learning. As we mentioned in the paper, we did not consider the recent advanced architectures, such as ResNet \cite{he2016identity} and DenseNet \cite{huang2016densely}.  

\begin{table}[h!]
  \caption{Supervised learning performance on CIFAR-10. Each setting is repeated for five times.}
  \label{cnn_mnist_sup}
  \centering
  \begin{tabular}{lcll}
    \hline

    Method     										& Error rate ($\%$) \\
    \hline 
    Network in Network \shortcite{lin2013network}				&   8.81     \\ 
    All-CNN \shortcite{springenberg2014striving}				&   7.25      \\ 
    Deep Supervised Net \shortcite{lee2015deeply}			&   7.97     \\ 
    Highway Network  \shortcite{srivastava2015highway}		&   7.72     \\ 
    $\Pi$ model \shortcite{laine2016temporal}     				&   5.56     \\ 
    Temportal ensembling \shortcite{laine2016temporal}      		&   5.60     \\ 
    VAT \shortcite{miyato2017virtual}     					&   5.81     \\ 
    \hline
    $\Pi$ model (our implementation)     					&   5.77 $\pm$ 0.11    \\
    VAT (our implementation)							&   5.65 $\pm$ 0.17    \\
    AdD											&   5.46 $\pm$ 0.16    \\
    VAdD (KL)										&   5.27 $\pm$ 0.10    \\
    VAdD (QE)										&   \textbf{5.24} $\pm$ \textbf{0.12}    \\
    \hline 
    VAdD (KL) + VAT									&   \textbf{4.40} $\pm$ \textbf{0.12}    \\
    VAdD (QE) + VAT									&   4.73 $\pm$ 0.04    \\
    \hline 
  \end{tabular}
\end{table}

\subsection{D.2. CIFAR-10 : Semi-supervised classification results with additional baselines}

We compared the reported performances of additional baseline models for the semi-supervised learning. Our implementation reproduced the closed performance from their reported results, and showed the performance improvement from adversarial dropout. 

\begin{table}[h!]
  \caption{Semi-supervised learning task on CIFAR-10 with 4,000 labeled examples. Each setting is repeated for five times.}
  \label{cnn_mnist_sup}
  \centering
  \begin{tabular}{ll}
    \hline

    Method     								& Error rate ($\%$) \\ \hline
    Ladder network \shortcite{rasmus2015semi} 				&  20.40     \\ 
    CatGAN \shortcite{springenberg2015unsupervised}     		&  19.58      \\ 
    GAN with feature matching \shortcite{DBLP:journals/corr/SalimansGZCRC16} &  18.63     \\ 
    $\Pi$ model \shortcite{laine2016temporal}     				&  12.36      \\ 
    Temportal ensembling \shortcite{laine2016temporal}     		&  12.16     \\ 
    Sajjadi et al. \shortcite{sajjadi2016regularization} 			&  11.29     \\ 
    VAT \shortcite{miyato2017virtual}     					&  10.55     \\ 
    \hline
    $\Pi$ model (our implementation)     					&   12.62  $\pm$ 0.29     \\
    VAT (our implementation)							&   11.96 $\pm$ 0.10    \\
    VAdD (KL)										&   11.68 $\pm$ 0.19    \\
    VAdD (QE)										&   \textbf{11.32} $\pm$ \textbf{0.11}    \\
   \hline
    VAdD (KL) + VAT									&   10.07 $\pm$ 0.11    \\
    VAdD (QE) + VAT									&   \textbf{9.22} $\pm$ \textbf{0.10}    \\
    \hline 
  \end{tabular}
\end{table}

\section{Appendix E. Proof of Linear Regression Regularization}

In this section, we showed the detailed proof of regularization terms from adversarial training and adversarial dropout.

\subsection{Linear Regression with Adversarial Training}  
Let $\mathbf{x}_i \in \mathbb{R}^{D}$ be a data point and $y_i \in \mathbb{R}$ be a target where $i=\{1,...,N\}$. The objective of linear regression is to find a $\mathbf{w} \in \mathbb{R}^{D}$ that minimizes $l(\mathbf{w})=\sum_i \| y_i - \mathbf{x}_i^T\mathbf{w} \|^{2}$. 

To express adversarial examples, we denote $\tilde{\mathbf{x}}_i = \mathbf{x}_i + \mathbf{r}_i^{adv}$ as the adversarial example of $\mathbf{x}_i$ where $\mathbf{r}_i^{adv}=\delta sign( \boldsymbol{\bigtriangledown}_{\mathbf{x}_i} l(\mathbf{w}))$ utilizing the fast gradient sign method (FGSM) \cite{goodfellow2014explaining}, $\delta$ is a controlling parameter representing the degree of adversarial noises. With the adversarial examples, the objective function of adversarial training can be viewed as follows:
\begin{gather}
l_{AT}(\mathbf{w})=\sum_i \| y_i - (\mathbf{x}_i+\mathbf{r}^{adv}_i)^T\mathbf{w} \|^{2}
\end{gather}
This can be divided to
\begin{gather}
l_{AT}(\mathbf{w})=l(\mathbf{w}) - 2 \sum_i (y_i - \mathbf{x}_i^T\mathbf{w})\mathbf{w}^T\mathbf{r}^{adv}_i \\
\: \: + \sum_i \mathbf{w}^T (\mathbf{r}^{adv}_i)^T (\mathbf{r}^{adv}_i) \mathbf{w} \nonumber
\end{gather}
where $l(\mathbf{w})$ is the loss function without adversarial noise. Note that the gradient is $\boldsymbol{\bigtriangledown}_{\mathbf{x}_i} l(\mathbf{w})= - 2  (y_i - \mathbf{x}_i^T\mathbf{w}) \mathbf{w}$, and $\mathbf{a}^Tsign(\mathbf{a})=\| \mathbf{a} \|_1$. The above equation can be transformed as the following:
\begin{gather}
l_{AT}(\mathbf{w})=l(\mathbf{w}) + \sum_{ij} |\delta \bigtriangledown_{x_{ij}} l(\mathbf{w})| + \delta^2 \mathbf{w}^T \Gamma_{AT} \mathbf{w},
\end{gather}
where $\Gamma_{AT}= \sum_i sign( \boldsymbol{\bigtriangledown}_{\mathbf{x}_i} l(\mathbf{w}))^T sign( \boldsymbol{\bigtriangledown}_{\mathbf{x}_i} l(\mathbf{w}) )$. 

\subsection{Linear Regression with Adversarial Dropout}  

To represent adversarial dropout, we denote $\tilde{\mathbf{x}}_i = \boldsymbol{\epsilon}_i^{adv}\odot\mathbf{x}_i$ as the adversarially dropped input of $\mathbf{x}_i$ where $\boldsymbol{\epsilon}_i^{adv}=argmax_{\boldsymbol{\epsilon}; \|\boldsymbol{\epsilon}_i - 1 \|_{2} \leq k} \| y_i - (\boldsymbol{\epsilon}_i \odot \mathbf{x}_i)^T\mathbf{w} \|^{2}$ with the hyper-parameter, $k$, controlling the degree of adversarial dropout. For simplification, we used one vector as the base condition of a adversarial dropout. If we applied our proposed algorithm, the adversarial dropout can be defined as follows:
\begin{gather}
\epsilon_{ij}^{adv} = \left\{ 
\begin{array}{rcl}
 0 &  & \mbox{if} \: x_{ij}\bigtriangledown_{x_{ij}} l(\mathbf{w}) \leq min\{s_{ik}, 0\}  \\
 1 &  & \mbox{otherwise} 
\end{array} \right.
\end{gather}
where $s_{ik}$ is the $k^{th}$ lowest element of $ \mathbf{x}_{i} \odot \boldsymbol{\bigtriangledown}_{\mathbf{x}_{i}} l(\mathbf{w})$. This solution satisfies the constraint, $\|\boldsymbol{\epsilon}_i - 1 \|_{2} \leq k$. With this adversarial dropout condition, the objective function of adversarial dropout can be defined as the following:
\begin{gather}
l_{AdD}(\mathbf{w})=\sum_i \| y_i - (\boldsymbol{\epsilon}_i^{adv} \odot \mathbf{x}_i)^T\mathbf{w} \|^{2} 
\end{gather}
This can be divided to 
\begin{gather}
l_{AdD}(\mathbf{w})=l(\mathbf{w}) + 2 \sum_i (y_i - \mathbf{x}_i^T\mathbf{w}) (( 1 - \boldsymbol{\epsilon}_i^{adv}) \odot \mathbf{x}_i)^T\mathbf{w} \\
\: \: + \sum_i \mathbf{w}^T (( 1 - \boldsymbol{\epsilon}_i^{adv}) \odot \mathbf{x}_i)^T (( 1 - \boldsymbol{\epsilon}_i^{adv}) \odot \mathbf{x}_i)\mathbf{w} \nonumber
\end{gather}
The second term of the right handside can be viewed as
\begin{gather}
2 \sum_i (y_i - \mathbf{x}_i^T\mathbf{w}) \sum_j ( 1 - \epsilon_{ij}^{adv})x_{ij} w_j.
\end{gather}
By defining a set $S_{i} = \{ j | \mathbf{\epsilon}_{ij}^{adv}=0 \}$, the second term can be transformed as the following.
\begin{gather}
 - \sum_i \sum_{j \in S_{i}} -2(y_i - \mathbf{x}_i^T\mathbf{w}) w_j x_{ij}.
\end{gather}
Note that the gradient is $\bigtriangledown_{x_{ij}} l(\mathbf{w})= - 2  (y_i - \mathbf{x}_i^T\mathbf{w}) \mathbf{w}_j$ and $x_{ij}\bigtriangledown_{x_{ij}} l(\mathbf{w})$ is always negative when $j \in S_{i}$. The second term can be re-defined as the following.
\begin{gather}
 \sum_i \sum_{j \in S_{i}} | x_{ij} \bigtriangledown_{x_{ij}} l(\mathbf{w})|
\end{gather}
Finally, the objective function of adversarial dropout is re-organized.
\begin{gather}
l_{AdD}(\mathbf{w})=l(\mathbf{w}) + \sum_{i} \sum_{j \in S_{i}} | x_{ij} \bigtriangledown_{x_{ij}} l(\mathbf{w})| +  \mathbf{w}^T \Gamma_{AdD} \mathbf{w},
\end{gather}
where $S_{i} = \{ j | \epsilon_{ij}^{adv}=0 \}$ and  $\Gamma_{AdD}=\sum_{i} (( 1 - \boldsymbol{\epsilon}_i^{adv}) \odot \mathbf{x}_i)^T (( 1 - \boldsymbol{\epsilon}_i^{adv}) \odot \mathbf{x}_i)$. 

\end{document}